\documentclass[10pt,twocolumn,letterpaper]{article}

\usepackage{cvpr}
\usepackage{times}
\usepackage{epsfig}
\usepackage{graphicx}
\usepackage{amsmath}
\usepackage{amssymb}
\usepackage{subfig}
\usepackage{amsmath}
\usepackage{hhline}
\usepackage{multirow}
\usepackage{array}
\newcolumntype{k}[1]{>{\centering\arraybackslash}m{#1}}
\newcolumntype{m}[1]{>{\centering\arraybackslash}p{#1}}
\usepackage{enumitem}

\usepackage{rotating}
\pdfoutput=1

\usepackage[pagebackref=true,breaklinks=true,letterpaper=true,colorlinks,bookmarks=false]{hyperref}

 \cvprfinalcopy 

\def\cvprPaperID{3694} 

\ifcvprfinal\fi
\begin{document}
\setlength{\belowdisplayskip}{3.8pt} 
\setlength{\belowdisplayshortskip}{3.8pt}
\setlength{\abovedisplayskip}{3.8pt} 
\setlength{\abovedisplayshortskip}{3.8pt}

\title{MagnifyMe: Aiding Cross Resolution Face Recognition via Identity Aware Synthesis }

\author{Maneet Singh, Shruti Nagpal, Richa Singh, Mayank Vatsa, and Angshul Majumdar \\
IIIT-Delhi, India \\
{\small \{maneets, shrutin, rsingh, mayank, angshul\}@iiitd.ac.in}
}

\maketitle

\begin{abstract}
Enhancing low resolution images via super-resolution or image synthesis for cross-resolution face recognition has been well studied. Several image processing and machine learning paradigms have been explored for addressing the same. In this research, we propose \textit{Synthesis via Deep Sparse Representation} algorithm for synthesizing a high resolution face image from a low resolution input image. The proposed algorithm learns multi-level sparse representation for both high and low resolution gallery images, along with an identity aware dictionary and a transformation function between the two representations for face identification scenarios. With low resolution test data as input, the high resolution test image is synthesized  using the identity aware dictionary and transformation which is then used for face recognition. The performance of the proposed SDSR algorithm is evaluated on four databases, including one real world dataset. Experimental results and comparison with existing seven algorithms demonstrate the efficacy of the proposed algorithm in terms of both face identification and image quality measures. 
\end{abstract}

\section{Introduction}



\begin{figure}[!t]
\centering
\includegraphics[width = 3.2in]{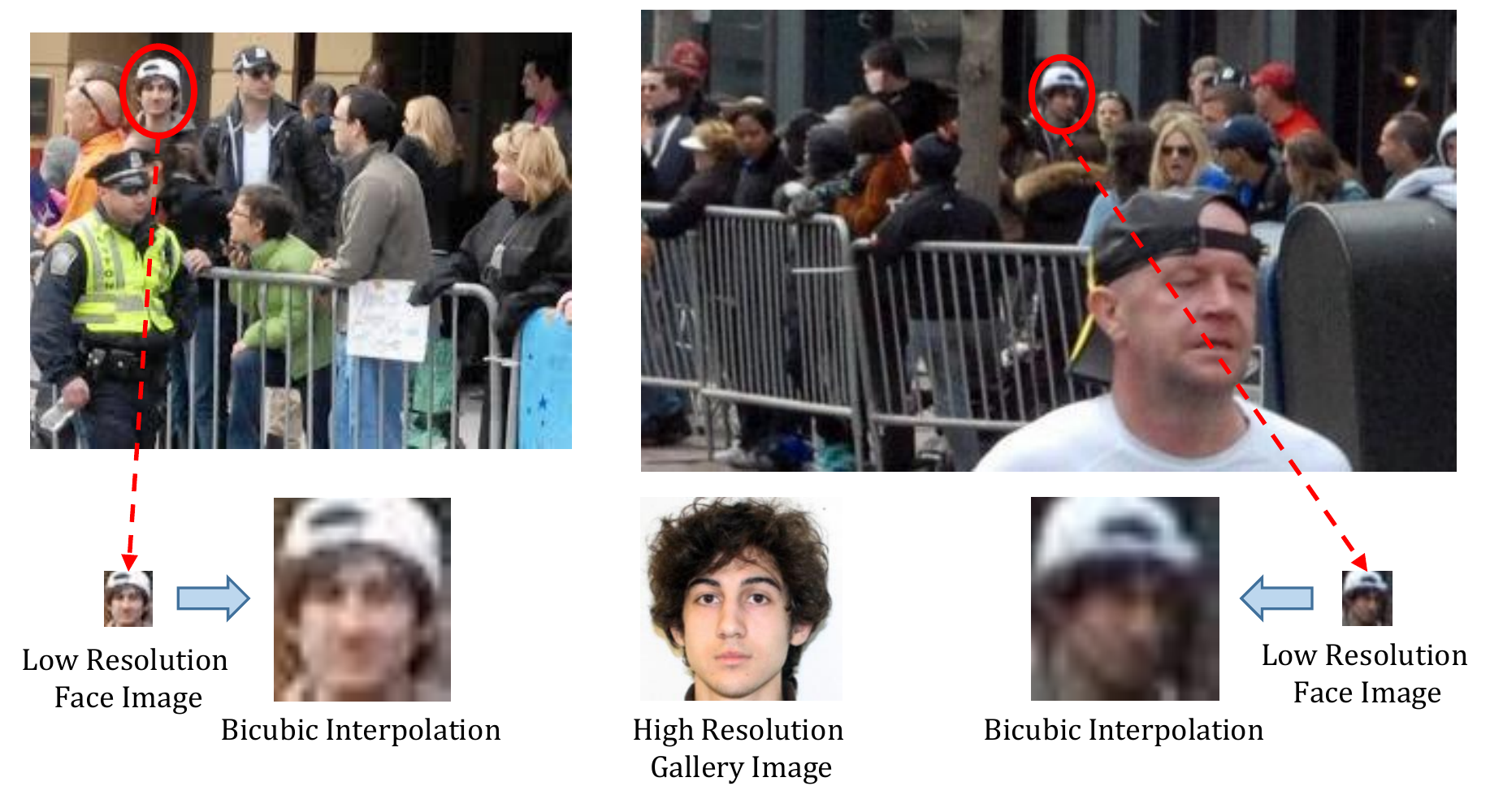}
\vspace{-10pt}
\caption{Images captured few minutes before Boston Marathon Bombing, 2013, of suspect Dzhokhar Tsarnaev (circled). The resolution of the circled image is less than $24\times24$, which is interpolated to ($96\times96$).}
\vspace{-15pt}
\label{intro}
\end{figure}

Group images are often captured from a distance, in order to capture multiple people in the image. In such cases, the resolution of each face image is relatively smaller, thereby resulting in errors during automated tagging. Similarly, in surveillance and monitoring applications, cameras are often designed to cover the maximum field of view, this often limits the size of face images captured, especially for individuals at a distance. If we use these images to match against high resolution images, e.g., profile images on social media or mugshot images captured by law enforcement, then resolution gap between the two may lead to incorrect results. This task of matching a low resolution input image against a database of high resolution images is referred to as cross resolution face recognition and it is a challenging covariate of face recognition with widespread applications.

Several researchers have shown that the performance of state-of-the-art (SOTA) algorithms reduces while matching cross-resolution face images \cite{himanshu, wang16CVPR, Wang2013}. In order to overcome this limitation, an intuitive approach is to generate a high resolution image for the given low resolution input, which can be provided as input to the face recognition engine. Figure \ref{intro} shows a sample real world image captured before the Boston Bombing (2013). Since the person of interest is at a distance, the face captured is thus of low resolution. Upon performing bicubic interpolation to obtain a high resolution image, results in an image suffering from blur and poor quality.  
With the ultimate aim of high recognition performance, the generated high resolution image should have good quality while preserving the identity of the subject. As elaborated in the next subsection, while there exist multiple synthesis or super resolution techniques, we hypothesize that utilizing a (domain) face-specific, recognition-oriented model for face synthesis should result in improved recognition performance, especially for close-set recognition scenarios. To this effect, this work presents a novel domain specific identity aware \textit{Synthesis via Deep Sparse Coding} algorithm for synthesizing a high resolution face image from a given low resolution input image.




\subsection{Literature Review}
In literature, different techniques have been proposed to address the problem of cross resolution face recognition. These can broadly be divided into transformation based techniques and non-transformation based techniques. Transformation based techniques address the resolution difference between images by explicitly introducing a transformation function either at the image or at the feature level. Non-transformation techniques propose to learn/extract resolution invariant features or classifiers, in order to address the resolution variations \cite{himanshu, wang16CVPR}. In 2013, Wang \textit{et al.} \cite{Wang2013} present an exhaustive review of the proposed techniques for addressing cross resolution face recognition. 

Peleg and Elad \cite{peleg} propose a statistical model that uses Minimum Mean Square Error estimator on high and low resolution image pair patches for prediction. Lam \cite{jian2015} propose a Singular Value Decomposition based approach for super resolving low resolution face images. Researchers have also explored the domain of representation learning to address the problem of cross resolution face recognition. Yang \textit{et al.} \cite{yang} propose learning dictionaries for low and high resolution image patches jointly followed by learning a mapping between the two. Yang \textit{et al.} \cite{icb} propose a Sparse Representation-based Classification approach in which the face recognition and hallucination constraints are solved simultaneously. Gu \textit{et al.} \cite{gu_iccv15} propose convolutional sparse coding where an image is divided into patches and filters are learned to decompose a low resolution image into features. A mapping is learned to predict high resolution feature maps from the low resolution features. Mundunuri and Biswas \cite{mundunuri16} propose a multi-dimensional scaling and stereo cost technique to learn a common transformation matrix for addressing the resolution variations. 

A parallel area of research is that of super-resolution, where research has focused on obtaining a high resolution image from a given low resolution image, with the objective of maintaining/improving the visual quality of the input \cite{baker2000, bayesianSR, wang14Rev, scdl}. There has been significant advancement in the field of super-resolution over the past several years including recent representation learning architectures \cite{pixel17Iccv, srcnn, fsrcnn, ledig17Cvpr, tong17Iccv} being proposed for the same. It is important to note that while such techniques can be utilized for addressing cross resolution face recognition, however, they are often not explicitly trained for face images, or for providing recognition-oriented results.

\subsection{Research Contributions}
 This research focuses on cross resolution face recognition by proposing a recognition-oriented image synthesis algorithm, capable of handling large magnification factors. We propose a deep sparse representation based transfer learning approach termed as Synthesis via Deep Sparse Representation (SDSR). The proposed identity aware synthesis algorithm can be incorporated as a pre-processing module prior to any existing face recognition engine to enhance the resolution of a given low resolution input. In order to ensure recognition-oriented synthesis, the proposed model is trained using a gallery database having a single image per subject. The results are demonstrated with four databases and the effectiveness is evaluated in terms of both no-reference image quality measure of the synthesized images and face identification accuracies with existing face recognition models.

\section{Synthesis via Deep Sparse Representation}
\label{algo}
Dictionary Learning algorithms have an inherent property of representing a given sample as a sparse combination of it's basis functions \cite{doubleSparsity}. This property is utilized in the proposed SDSR algorithm to synthesize a high resolution image from a given low resolution input. The proposed model learns a transformation between the representations of low and high resolution images. That is, instead of interpolating the pixel values, this work focuses on interpolating a more abstract representation. Further, motivated by the abstraction capabilities of deep learning, we propose to learn the transformation from deeper levels of representation. Unlike traditional dictionary learning algorithms, we propose to learn the transformation at deeper levels of representation. This leads to the key contribution of this work: \textit{Synthesis via Deep Sparse Representation (SDSR)}, a transfer learning approach for synthesizing a high resolution image for a given low resolution input. 

\subsection{Preliminaries}
Let $\mathbf{X = [x^\textit{1}| x^\textit{2}| ... |x^\textit{n}]}$ be the input training data with $n$ samples. Dictionary learning algorithms learn a \textit{dictionary} ($\mathbf{D}$) and \textit{sparse representations} ($\mathbf{A}$) using data ($\mathbf{X}$). The objective function of dictionary learning is written as:
\begin{equation} \label{eqDict}
\min_{\textit{$\mathbf{D, A}$}}  \frac{1}{n} \sum \limits_{i=1}^{n} \Big( \frac{1}{2} \left \| \mathbf{x^{\textit{i}} - D \boldsymbol{\alpha^\textit{i}}} \right \|_{F}^{2} + \lambda \left\| \boldsymbol{\alpha^\textit{i}}\right\|_1 \Big)
\end{equation} 
\noindent where, $\mathbf{A = \boldsymbol{[\alpha^\textit{1}| \alpha^\textit{2}| .... |\alpha^\textit{n}]}}$ are the sparse codes, $\left\| \cdot \right\|_1$ represents $\ell_1$-norm, and $\lambda$ is the regularizing constant that controls how much weight is given to induce sparsity in the representations. In Eq. \ref{eqDict}, the first term minimizes the reconstruction error of the training samples, and the second term is a regularization term on the sparse codes. 

In literature, researchers have proposed extending a single level dictionary to a multi-level dictionary to learn multiple levels of representations of the given data \cite{sparseFilt, deepDict, multiLevel}. 
A $k-$level deep dictionary learns $k$ dictionaries $\mathbf{D = \{D^\textit{1}, ..., D^\textit{k}\}}$ and sparse coefficients $\mathbf{A = \{A^\textit{1}, ..., A^\textit{k}\}}$ for a given input $\mathbf{X}$. 
\begin{equation} \label{eqDeepDict}
\min_{\textit{$\mathbf{D, A}$}} \  \frac{1}{n} \sum \limits_{i=1}^{n}\Big( \frac{1}{2} \left \| \mathbf{x^\textit{i} - D^\textit{1}...D^\textit{k} \boldsymbol{\alpha^\textit{k,i}}} \right \|_{F}^{2} + \lambda \left\| \boldsymbol{\alpha^\textit{k,i}}\right\|_1 \Big)
\end{equation} 
The architecture of deep dictionary is inspired from the deep learning techniques where deeper layers of feature learning enhance the level of abstraction learned by the network, thereby learning meaningful latent variables. 

\begin{figure*}
\centering
\includegraphics[width= 6.9in]{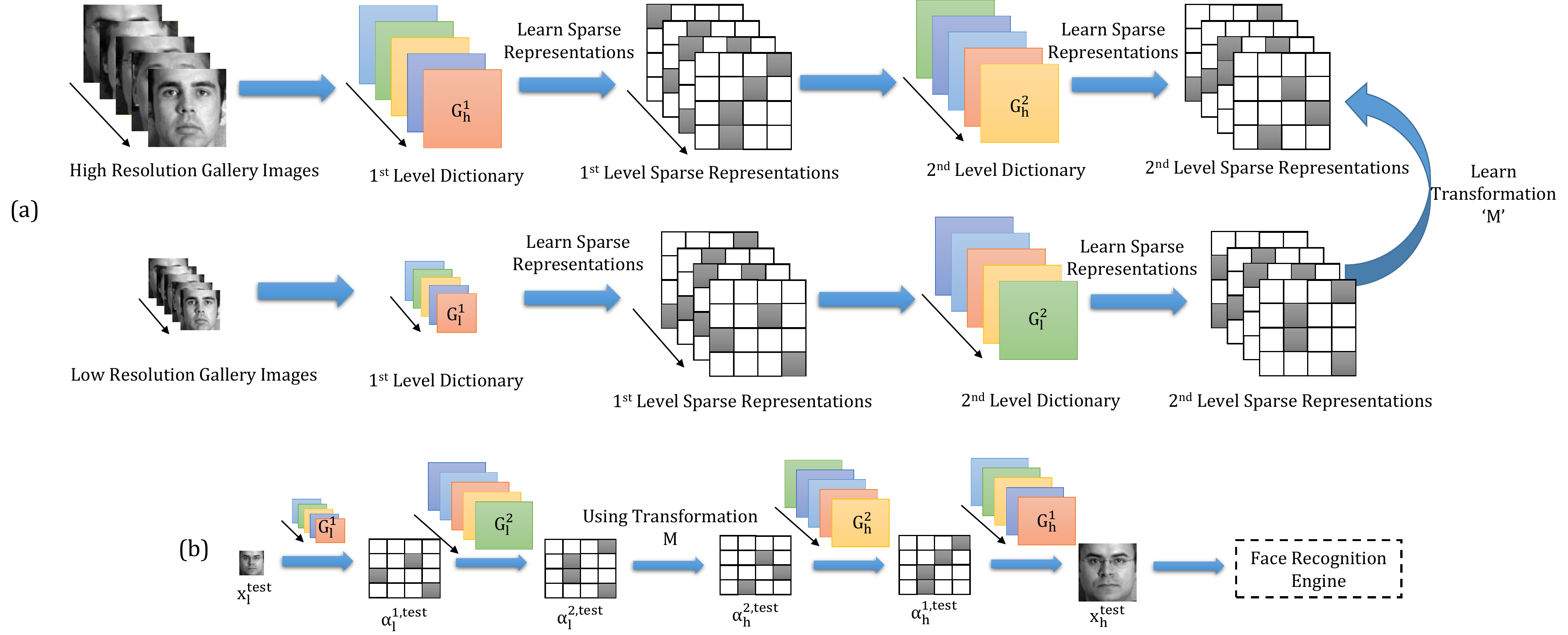} 
\vspace{-5pt}
\caption{Synthesis via Deep Sparse Representation algorithm for 2-level deep dictionary. (a) refers to the training of the model, while (b) illustrates the high resolution synthesis of a low resolution input. }
\vspace{-10pt}
\label{algo}
\end{figure*}

\subsection{SDSR Algorithm}
In real world scenarios of surveillance or image tagging, the task is to match the low resolution test images (probe) to the database of high resolution images known as \textit{gallery images}. Without loss of generality, we assume that the target comprises of high resolution gallery images while the source domain consists of low resolution images. In the proposed model, for low resolution face images $\mathbf{X}_{l}$, and high resolution face images $\mathbf{X}_{h}$, $k$ level deep dictionaries are learned in both source ($\mathbf{G_{L} = \{G_\textit{l}^\textit{1}, ...,  G_\textit{l}^\textit{k}\}}$) and target domain ($\mathbf{G_{H} = \{G_\textit{h}^\textit{1}, ..., G_\textit{h}^\textit{k}}\}$). It is important to note that the dictionaries are generated using the pre-acquired gallery images. Corresponding sparse representations, $\mathbf{A_{L} = \{A_\textit{l}^\textit{1}, ...,  A_\textit{l}^\textit{k}\}}$ and $\mathbf{A_{H} = \{A_\textit{h}^\textit{1}, ...,  A_\textit{h}^\textit{k}\}}$ are also learned for all $k$ levels, where $\boldsymbol{\mathbf{A_\textit{h}^\textit{k}} = [\alpha_\textit{h}^\textit{k,1}| \alpha_\textit{h}^\textit{k,2}| ... |\alpha_\textit{h}^\textit{k,n}]}$ are the representations learnt corresponding to the high resolution deep dictionary and $\boldsymbol{\mathbf{A_\textit{l}^\textit{k}} = [\alpha_\textit{l}^\textit{k,1}| \alpha_\textit{l}^\textit{k,2}| ... |\alpha_\textit{l}^\textit{k,n}]}$ are the representations learnt from the $k^{th}$ level dictionary for the low resolution images. The proposed algorithm learns a transformation, $\mathbf{M}$, between $\mathbf{A^\textit{k}_\textit{h}}$ and $\mathbf{A^\textit{k}_\textit{l}}$. The optimization formulation for Synthesis via Deep Sparse Representation (SDSR), a $k$-level deep dictionary is written as: 

\begin{equation} \label{eqProposed}
\begin{split}
\min_{\textit{$\substack{ \mathbf{G_{H}, A_{H},} \\ \mathbf{G_{L}, A_{L}, M} }$}} &  \frac{1}{n} \sum \limits_{i=1}^{n} \Big( \frac{1}{2} \left \| \mathbf{x_\textit{h}^\textit{i} - G_\textit{h}^\textit{1} ... G_\textit{h}^\textit{k} \boldsymbol{\alpha_\textit{h}^\textit{k,i} }} \right \|_{F}^\textit{2} + \sum \limits_{j = 1} ^{k} (\lambda_j \left\| \boldsymbol{\alpha_h^\textit{j,i}}\right\|_1)\\
& + \frac{1}{2} \left \| \mathbf{x_\textit{l}^\textit{i} - G_\textit{l}^\textit{1} ... G_\textit{l}^\textit{k} \boldsymbol{\alpha_\textit{l}^\textit{k,i}} }\right \|_{F}^{2} + \sum \limits_{j = 1} ^{k} (\lambda_j \left\| \boldsymbol{\alpha_l^\textit{j,i}}\right\|_1)\\
& + \lambda_M  \left\| \mathbf{\boldsymbol{\alpha_\textit{h}^\textit{k,i}} - M\boldsymbol{\alpha_\textit{l}^\textit{k,i}}} \right\|_F^2 \Big)
\end{split}
\end{equation}
where, $\lambda_j$ are regularization parameters which control the amount of sparsity in the learned representations of the $j^{th}$ layer, while $\lambda_M$ is the regularization constant for learning the transformation function. $\mathbf{G_{H}}$ and $\mathbf{G_{L}}$ correspond to the deep dictionaries learned for the high and low resolution gallery images, respectively. The SDSR algorithm learns multiple levels of dictionaries and corresponding representations for low and high resolution face images, along with a transformation between the features learned at the deepest layer.

\subsubsection{Training SDSR Algorithm}
Without loss of generality, training of the proposed SDSR algorithm is explained with $k = 2$ (shown in Figure \ref{algo}). For a two level deep dictionary, Eq. \ref{eqProposed} can be written as:
\begin{equation} \label{eqComplete}
\begin{gathered}
\min_{\textit{$\substack{ \mathbf{G_{H}, G_{L}} \\ \mathbf{A_{H}, A_{L}, M} }$}}   \frac{1}{n} \sum \limits_{i=1}^{n} \Big( \frac{1}{2} \left \| \mathbf{x_\textit{h}^\textit{i} - G_\textit{h}^\textit{1}G_\textit{h}^\textit{2} \boldsymbol{\alpha_\textit{h}^\textit{2,i}}} \right \|_{F}^\textit{2} + \lambda_1 \left\| \boldsymbol{\alpha_\textit{h}^\textit{1,i }}\right\|_1 \\
+ \lambda_2 \left\| \boldsymbol{\alpha_\textit{h}^\textit{2,i }}\right\|_1 + \frac{1}{2} \left \| \mathbf{x_\textit{l}^\textit{i} - G_\textit{l}^\textit{1}G_\textit{l}^\textit{2} \boldsymbol{\alpha_\textit{l}^\textit{2,i}} }\right \|_{F}^\textit{2} + \lambda_1 \left\| \boldsymbol{\alpha_\textit{l}^\textit{1,i }}\right\|_1 \\
+ \lambda_2 \left\| \boldsymbol{\alpha_\textit{l}^\textit{2,i }}\right\|_1 + \lambda_M  \left\| \mathbf{\boldsymbol{\alpha_\textit{h}^\textit{2,i} - \mathbf{M}\alpha_\textit{l}^\textit{2,i}}} \right\|_F^2 \Big)
\end{gathered}
\end{equation}
Since the number of variables in Eq. \ref{eqComplete} is large (even more for deeper dictionaries), directly solving the optimization problem may provide incorrect estimates and lead to overfitting. Therefore, greedy layer by layer training is applied. It is important to note that since there is a $l_{1}$-norm regularizer on the coefficients of the first and the second layer, the dictionaries $\mathbf{G}_h^1$ and $\mathbf{G}_h^2$ cannot be collapsed into one dictionary. In order to learn the estimates, Eq. \ref{eqComplete} is split into learning the first level representation, second level representation, and the transformation. From Eq. \ref{eqDeepDict}, the optimization function for two level deep dictionary is as follows: 
\begin{equation} \label{eqTarget}
\min_{\textit{$\substack{\mathbf{G^\textit{1},D^\textit{2},} \\ \mathbf{A^\textit{1}, A^\textit{2}}}$}}  \frac{1}{n} \sum \limits_{i=1}^{n} \Big( \frac{1}{2} \left \| \mathbf{x^\textit{i} - G^\textit{1}G^\textit{2} \boldsymbol{\alpha^\textit{2,i}} }\right \|_{F}^\textit{2} + \lambda_1 \left\| \boldsymbol{\alpha^\textit{1,i}} \right\|_1 +  \lambda_2 \left\| \boldsymbol{\alpha^\textit{2,i}} \right\|_1 \Big)
\end{equation} 
\indent Assuming an intermediate variable, $\boldsymbol{\alpha^\textit{1,i} = }\mathbf{G^\textit{2}\boldsymbol{\alpha^\textit{2,i}}}$ such that $\mathbf{A^\textit{1} = \boldsymbol{[\alpha^\textit{1,1}| \alpha^\textit{1,2}| ... |\alpha^\textit{1,n}]}}$, the above equation can be modeled as a step-wise optimization of the following two equations:
\begin{equation} \label{eqTargetGreedyFirst}
\min_{\textit{$\mathbf{G^\textit{1}, A^\textit{1}}$}}  \frac{1}{n} \sum \limits_{i=1}^{n} \Big( \frac{1}{2} \left \| \mathbf{x^\textit{i} - G^\textit{1}\boldsymbol{\alpha^\textit{1,i} }}\right \|_{F}^{2} + \lambda_1 \left\| \boldsymbol{\alpha^\textit{1,i}} \right\|_1 \Big)
\end{equation}
\begin{equation} \label{eqTargetGreedySecond}
\min_{\textit{$\mathbf{G^\textit{2}, A^\textit{2}}$}}  \frac{1}{n} \sum \limits_{i=1}^{n} \Big( \frac{1}{2} \left \|\boldsymbol{\alpha^\textit{1,i}} - \mathbf{G^\textit{2} \boldsymbol{\alpha^\textit{2,i}} }\right \|_{F}^{2} + \lambda_2 \left\| \boldsymbol{\alpha^\textit{2,i} }\right\|_1 \Big)
\end{equation}
\indent A deep dictionary of two levels (Eq. \ref{eqTarget}) requires two steps for learning (Eq. \ref{eqTargetGreedyFirst}, \ref{eqTargetGreedySecond}). Upon extending the formulation to $k$ level deep dictionary, it would require exactly $k$ steps for optimization. The proposed SDSR algorithm (Eq. \ref{eqProposed}) builds upon the above and utilizes $k+1$ steps based greedy layer-wise learning for a $k$ level deep dictionary. $k$ steps are for learning representations using the deep dictionary architecture and the $k+1^{th}$ step is for learning the transformation between the final representations. Therefore, Eq. \ref{eqComplete} is solved using an independent three step approach:
\textbf{(i)} learn first level source (low resolution) and target (high resolution) domain dictionaries,
\textbf{(ii)} learn second level low and high resolution image dictionaries, and 
\textbf{(iii)} learn a transformation between the final representations.

Using the concept in Eq. \ref{eqTarget} - \ref{eqTargetGreedySecond}, in the \textbf{first step}, two separate level-1 (i.e. $k=1$) dictionaries are learned from the given input data for the low resolution ($\mathbf{G^\textit{1}_\textit{l}}$) and high resolution ($\mathbf{G_\textit{h}^\textit{1}}$) face images independently. Given the training data consisting of low ($\mathbf{X_\textit{l}}$) and high ($\mathbf{X_\textit{h}}$) resolution face images, the following minimization is applied for the two domains respectively:
\begin{equation} \label{eqDictSource}
\min_{\textit{$\mathbf{G_\textit{l}^\textit{1},A_\textit{l}^\textit{1}}$}} \frac{1}{n} \sum \limits_{i=1}^{n}  \frac{1}{2} \left \| \mathbf{x_\textit{l}^\textit{i} - G_\textit{l}^\textit{1} \boldsymbol{\alpha_\textit{l}^\textit{1,i} }} \right \|_{2}^{2}  +  \lambda_1 \left\| \boldsymbol{\alpha_\textit{l}^\textit{1,i}} \right\|_1 
\end{equation}
\begin{equation} \label{eqDictTarget}
\min_{\textit{$\mathbf{G_\textit{h}^\textit{1},A_\textit{h}^\textit{1}}$}} \frac{1}{n} \sum \limits_{i=1}^{n}  \frac{1}{2} \left \|\mathbf{x_\textit{h}^\textit{i} - G_\textit{h}^\textit{1} \boldsymbol{\alpha_\textit{h}^\textit{1,i}}} \right \|_{2}^{2}  +  \lambda_1 \left\| \boldsymbol{\alpha_\textit{h}^\textit{1,i}} \right\|_1 
\end{equation}
\noindent here, $\mathbf{A_\textit{l}^\textit{1} = \boldsymbol{[\alpha_\textit{l}^\textit{1,1}| \alpha_\textit{l}^\textit{1,2}| ... |\alpha_\textit{l}^\textit{1,n}]}}$ and $\mathbf{A_\textit{h}^\textit{1} = \boldsymbol{[\alpha_\textit{h}^\textit{1,1}| \alpha_\textit{h}^\textit{1,2}| ... |\alpha_\textit{h}^\textit{1,n}]}}$ refer to the level-1 sparse codes learned for the low and high resolution images respectively. Each of the above two equations can be optimized independently using an alternating minimization dictionary learning technique over the dictionary $\mathbf{G}^1$ and representation $\mathbf{A}^1$ \cite{dictLearning}. After this step, $\mathbf{G_\textit{h}^\textit{1}, G_\textit{l}^\textit{1}}$ (dictionaries) and $\mathbf{A_\textit{h}^\textit{1}, A_\textit{l}^\textit{1}}$ (representations) are obtained for the two varying resolution data. 

In the \textbf{second step}, a deep dictionary is created by learning the second level dictionaries ($\mathbf{G_\textit{l}^\textit{2}}, \mathbf{G_\textit{h}^\textit{2}}$) using the representations obtained from the first level ($\mathbf{A_\textit{h}^\textit{1}}$ and $\mathbf{A_\textit{l}^\textit{1}}$). That is, two separate dictionaries, one for low resolution images and one for high resolution images are learned using the representations obtained at the first level as input features. The equations for this step can be written as follows:
\begin{equation} \label{eqAlphaDictSource}
\min_{\textit{$\mathbf{G_\textit{l}^\textit{2} ,\boldsymbol{A_\textit{l}^\textit{2}}}$}} \frac{1}{n} \sum \limits_{i=1}^{n}  \frac{1}{2} \left \|\boldsymbol{\alpha_\textit{l}^\textit{1,i}} - \mathbf{G_\textit{l}^\textit{2} \boldsymbol{\alpha_\textit{l}^\textit{2,i}}} \right \|_{2}^{2}  +  \lambda_2 \left\| \boldsymbol{\alpha_\textit{l}^\textit{2,i}} \right\|_1 
\end{equation}
\begin{equation} \label{eqAlphaDictTarget}
\min_{\textit{$\mathbf{G_\textit{h}^\textit{2},\boldsymbol{A_\textit{h}^\textit{2}}}$}} \frac{1}{n} \sum \limits_{i=1}^{n}  \frac{1}{2} \left \| \boldsymbol{\alpha_\textit{h}^\textit{1,i}} - \mathbf{G_\textit{h}^\textit{2}  \boldsymbol{\alpha_\textit{h}^\textit{2,i}}} \right \|_{2}^{2}  +  \lambda_2 \left\| \boldsymbol{\alpha_\textit{h}^\textit{2,i}} \right\|_1 
\end{equation}
\noindent here, $\mathbf{A_\textit{l}^\textit{2} = \boldsymbol{[\alpha_\textit{l}^\textit{2,1}| \alpha_\textit{l}^\textit{2,2}| ... |\alpha_\textit{l}^\textit{2,n}]}}$ is the final representation obtained for the low resolution images and $\mathbf{A_\textit{h}^\textit{2} = \boldsymbol{[\alpha_\textit{h}^\textit{2,1}| \alpha_\textit{h}^\textit{2,2}| ... |\alpha_\textit{h}^\textit{2,n}]}}$ refers to the representation obtained for the high resolution images. Similar to the previous step, the equations can be solved independently using alternating minimization over the dictionary and representations. After this step, $\mathbf{G_\textit{l}^\textit{2}, G_\textit{h}^\textit{2}, A_\textit{l}^\textit{2}}$ and $\mathbf{A_\textit{h}^\textit{2}}$ are obtained. 

In order to synthesize from one resolution to another, the \textbf{third step} of the algorithm involves learning a transformation between the deep representations of the two resolutions (i.e. $\mathbf{A_\textit{l}^\textit{2}}$ and $\mathbf{A_\textit{h}^\textit{2}}$). The following minimization is solved to obtain a transformation $\mathbf{M}$.
\begin{equation} \label{eqMapping}
\min_{\textit{$\mathbf{M}$}} \frac{1}{2} \left \|\mathbf{A_\textit{h}^\textit{2} - MA_\textit{l}^\textit{2}}\right \|_{F}^{2}  
\end{equation}
The above equation is a least square problem having a closed form solution. After training, the dictionaries ($\mathbf{G_\textit{l}^1, G_\textit{h}^1, G_\textit{l}^2, G_\textit{h}^2}$) and the transformation function ($\mathbf{M}$) are obtained which are then used at test time. 
\begin{table*}
\centering
\small
\caption{Summarizing the characteristics of the training and testing partitions of the databases used in experiments.}
\vspace{-10pt}
\label{tab:dbase}
\begin{tabular}{|l|m{4em}|m{4em}|m{4em}|m{4em}|m{6em}|m{11.5em}|}
\hline
\multirow{2}{*}{\textbf{Dataset}} & \textbf{Training Subjects} & \textbf{Training Images} &\textbf{Testing Subjects} & \textbf{Testing Images} & \textbf{Gallery Resolution}  & \multirow{2}{*}{\textbf{Probe Resolutions}}  \\ \hline

\textbf{CMU Multi-PIE \cite{mpie}} & 100 & 200 & 237 & 474 & \multirow{5}{*}{$96\times96$} & \\\cline{1-5}
\textbf{CAS-PEAL \cite{caspeal}} &500 & 659 & 540 & 705 & &\multirow{2}{*}{$8\times8$, $16\times16$, $24\times24$,}  \\\cline{1-5}
\textbf{Real World Scenarios \cite{himanshu}} & - & - & 1207 & 1222 &  & \\\cline{1-5} \cline{7-7}
\textbf{SCface \cite{scface}} & 50 & 300 & 80 & 480 & & $24\times24$, $32\times32$, $48\times48$ \\ \hline
\end{tabular}
\vspace{-5pt}
\end{table*}

\begin{table*}[]
\centering
\small
\caption{Rank-1 identification accuracies (\%) obtained using Verilook (COTS-I) for cross resolution face recognition. The target resolution is $96\times96$. The algorithms which do not support the required magnification factor are presented as $'-'$.}
\vspace{-10pt}
\label{verilookResults}
\begin{tabular}{|m{1.5em}|m{5em}|m{3.7em}|m{3.5em}|m{3.5em}|m{3.5em}|m{3.5em}|m{3.5em}|m{3.5em}|m{3.5em}|m{4em}|}
\hline
\multirow{2}{*}{} & \textbf{Probe Resolution} & \textbf{Original Image} & \textbf{Bicubic Interp.} & \textbf{Dong \textit{et al.} \cite{fsrcnn}} & \textbf{Kim \textit{et al.} \cite{kwan}} & \textbf{Gu \textit{et al.} \cite{gu_iccv15}} & \textbf{Dong \textit{et al.} \cite{srcnn}} &  \textbf{Peleg \textit{et al.} \cite{peleg}} & \textbf{Yang \textit{et al.} \cite{yang}} & \textbf{Proposed SDSR} \\ 
\hhline {|=|=|=|=|=|=|=|=|=|=|=|}

\multirow{5}{*}{\rotatebox[origin=c]{90}{\parbox[c]{2cm}{\centering \textbf{CMU MultiPIE}}} } &
8$\times$8 & 0.0$\pm$0.0 & 0.1$\pm$0.0 & - & - & - & - & - & - & \textbf{82.6$\pm$1.5} \\ \cline{2-11}
& 16$\times$16 & 0.0$\pm$0.0 & 1.1$\pm$0.1 & - & - & - & - & - & - & \textbf{91.1$\pm$1.3} \\ \cline{2-11}
& 24$\times$24 & 1.2$\pm$0.4 & 3.1$\pm$0.6 & 2.0$\pm$3.5 & 4.1$\pm$1.0 & 4.3$\pm$1.0 &  4.2$\pm$0.6 & - & - & \textbf{91.8$\pm$1.8} \\ \cline{2-11}
 & 32$\times$32 & 3.4$\pm$0.6 & 16.9$\pm$1.3 & 9.7$\pm$1.1& 17.5$\pm$1.1& 15.4$\pm$1.1 & 6.9$\pm$0.2 & 8.3$\pm$0.9 & - & \textbf{91.9$\pm$1.7} \\ \cline{2-11}
& 48$\times$48 & 91.9$\pm$1.1 & 95.8$\pm$0.4 & 85.8$\pm$0.7 & \textbf{96.2$\pm$0.6} & 93.1$\pm$0.9 & 95.5$\pm$1.1 & 92.8$\pm$0.4 & 94.0$\pm$0.6  & 91.5$\pm$1.5 \\ 
\hhline {|=|=|=|=|=|=|=|=|=|=|=|}

\multirow{5}{*}{\begin{turn}{90}\textbf{CAS-PEAL}\end{turn}} &
8$\times$8 & 0.0$\pm$0.0 & 0.0$\pm$0.0 & - & - & - & - & - & - & \textbf{92.8$\pm$0.7} \\ \cline{2-11}
& 16$\times$16 & 0.0$\pm$0.0 & 0.2$\pm$0.3 & - & - & - & - & - & - & \textbf{94.4$\pm$1.1} \\ \cline{2-11}
& 24$\times$24 & 0.4$\pm$0.6 & 14.9$\pm$1.7 & 0.4$\pm$0.2 & 2.3$\pm$0.8 & 1.9$\pm$0.7 & 2.5$\pm$0.7 & - & - & \textbf{95.3$\pm$1.4} \\ \cline{2-11}
& 32$\times$32 & 3.7$\pm$0.7 & 76.5$\pm$1.8 & 5.4$\pm$1.2 & 11.8$\pm$1.1 & 8.1$\pm$2.3 & 2.1$\pm$0.7 & 3.1$\pm$1.6 & - & \textbf{95.6$\pm$1.1} \\ \cline{2-11}
& 48$\times$48 & 63.4$\pm$1.7 & 90.8$\pm$1.5 & 46.5$\pm$2.5 & 75.8$\pm$2.3 & 77.7$\pm$2.1 & 72.0$\pm$0.7 & 74.0$\pm$2.6 & 73.3$\pm$3.3 & \textbf{95.4$\pm$1.5} \\ \cline{2-11}
\hhline {|=|=|=|=|=|=|=|=|=|=|=|}

\multirow{3}{*}{\begin{turn}{90}\textbf{SCface}\end{turn}} &
24$\times$24 & 1.1$\pm$0.2 & 0.8$\pm$0.1& 0.4$\pm$0.2 & 0.4$\pm$0.2 & 1.5$\pm$0.3 & 1.3$\pm$0.3 & - & - & \textbf{14.7$\pm$3.3} \\ \cline{2-11}
& 32$\times$32 & 1.8$\pm$0.5 & 2.5$\pm$0.3 & 2.2$\pm$0.4& 2.0$\pm$0.0 & 2.3$\pm$0.3 & 0.7$\pm$0.3 & 1.8$\pm$0.5 & - & \textbf{15.6$\pm$1.3} \\ \cline{2-11}
& 48$\times$48 & 6.5$\pm$0.6 & 9.5$\pm$1.9 & 6.9$\pm$0.6& 6.7$\pm$1.2 & 7.7$\pm$0.6 & 7.5$\pm$1.3 & 7.3$\pm$0.9 & 6.8$\pm$0.7 & \textbf{18.5$\pm$2.6} \\ \hline
\end{tabular}
\end{table*}

\vspace{6pt}
\subsubsection{Testing: Synthesizing High Resolution Face Image from Low Resolution Image}
During testing, a low resolution test image, $\mathbf{x_\textit{l}^\textit{test}}$, is input to the algorithm. 
Using the trained gallery based dictionaries, $\mathbf{G_\textit{l}^\textit{1}}$ and $\mathbf{G_\textit{l}^\textit{2}}$, first and second level representations ($\boldsymbol{\alpha_\textit{l}^\textit{1,test}, \alpha_\textit{l}^\textit{2,test}}$) are obtained for the given image:
\begin{equation} \label{eqAlphaSource}
\mathbf{x_\textit{l}^\textit{test} = G_\textit{l}^\textit{1}\boldsymbol{\alpha_\textit{l}^\textit{1,test}}}; \  \boldsymbol{\alpha_\textit{l}^\textit{1,test}} = \mathbf{G_\textit{l}^\textit{2}\boldsymbol{\alpha_\textit{l}^\textit{2,test}}}
\end{equation}
The transformation function, $\mathbf{M}$, learned in Eq. \ref{eqMapping}, is then used to obtain the second level high resolution representation ($\boldsymbol{\alpha_\textit{h}^\textit{2,test}}$).
\begin{equation} \label{eqMappedAlphaAlpha}
\mathbf{\boldsymbol{\alpha_\textit{h}^\textit{2,test}} = M\boldsymbol{\alpha_\textit{l}^\textit{2,test}}}
\end{equation}
Using Eq. \ref{eqDictTarget} and Eq. \ref{eqAlphaDictTarget} and the second level representation for the given image in the target domain, a \textit{synthesized} output of the given image is obtained. First $\boldsymbol{\alpha_\textit{h}^\textit{1,test}}$ is calculated with the help of $\mathbf{G_\textit{h}^\textit{2}}$ and then $\mathbf{x_\textit{h}^\textit{test}}$ is obtained using $\mathbf{G_\textit{h}^\textit{1}}$, which is the synthesized image in the target domain. 
\begin{equation} \label{eqAlphaTarget}
\boldsymbol{\alpha_\textit{h}^\textit{1,test}} = \mathbf{G_\textit{h}^2\boldsymbol{\alpha_\textit{h}^\textit{2,test}}}; \ \mathbf{x_\textit{h}^\textit{test} = G_\textit{h}^1\boldsymbol{\alpha_\textit{h}^\textit{1,test}}}
\end{equation}
It is important to note that the synthesized high resolution image is a sparse combination of the basis functions of the learned high resolution dictionary. In order to obtain a good quality, identity-preserving high resolution synthesis, the dictionary is trained with the pre-acquired high resolution database. This ensures that the basis functions of the trained dictionaries span the latent space of the images. As will be demonstrated via experiments as well, a key highlight of this algorithm is to learn good quality, representative dictionaries with a single sample per subject as well. The high resolution synthesized output image $\mathbf{x_\textit{h}^\textit{test}}$  can then be used by any face identification engine for recognition. 

\section{Databases and Experimental Protocol}
\label{sec:dbase}

The effectiveness of the proposed SDSR algorithm is demonstrated by evaluating the face recognition performance with original and synthesized images. Two commercial-off-the-shelf face recognition systems (COTS), Verilook (COTS-I) \cite{verilook} and Luxand (COTS-II) \cite{luxand} are used on four different face databases. For Verilook, the face quality and confidence thresholds are set to minimum, in order to reduce enrollment errors. The performance of the proposed algorithm is compared with six recently proposed super-resolution and synthesis techniques by Kim \textit{et al.} \cite{kwan}\footnote{https://people.mpi-inf.mpg.de/~kkim/supres/supres.htm} (kernel ridge regression), Peleg \textit{et al.} \cite{peleg}\footnote{http://www.cs.technion.ac.il/\~elad/software/} (sparse representation based statistical prediction model), Gu \textit{et al.} \cite{gu_iccv15}\footnote{http://www4.comp.polyu.edu.hk/~cslzhang/} (convolutional sparse coding), Yang \textit{et al.} \cite{yang}\footnote{http://www.ifp.illinois.edu/\~jyang29/} (dictionary learning), Dong \textit{et al.} \cite{srcnn}\footnote{http://mmlab.ie.cuhk.edu.hk/projects/SRCNN.html} (deep convolutional networks), and Dong \textit{et al.} \cite{fsrcnn}\footnote{http://mmlab.ie.cuhk.edu.hk/projects/FSRCNN.html} (deep convolutional networks) along with one of the most popular technique, bicubic interpolation. The results of the existing super-resolution algorithms are computed by using the models provided by the authors at the links provided in the footnotes. It is to be noted that not all the algorithms support all the levels of magnification. For instance, the algorithm proposed by Kim \textit{et al.} \cite{kwan} supports up to $4$ levels of magnification whereas, Yang \textit{et al.}'s algorithm supports up to 2 levels of magnification.

\textbf{Face Databases:} Table \ref{tab:dbase} summarizes the statistics of the databases in terms of training and testing partitions, along with the resolutions. Details of the databases are provided below: \\ 
\textbf{1. CMU Multi-PIE Dataset \cite{mpie}:} Images pertaining to $337$ subjects are selected with frontal pose, uniform illumination, and neutral expression. $100$ subjects are used for training while the remaining $237$ are in the test set. \\
\textbf{2. CAS-PEAL Dataset \cite{caspeal}} consists of face images of $1040$ subjects. All subjects have a single, high-resolution \textit{normal} image and the dataset contains images of different covariates such as lighting, expression, and distance. For this research, \textit{normal} images are used as the high resolution gallery database while face images under the \textit{distance} covariate are downsampled and used as probe images. \\
\textbf{3. SCface Dataset \cite{scface}: }It consists of $130$ subjects, each having one high resolution frontal face image and multiple low resolution images, captured from three distances using surveillance cameras. \\
\textbf{4. Real World Scenarios Dataset \cite{himanshu}} contains images of seven subjects associated with the London Bombing, Boston Bombing, and Mumbai Attacks. Each subject has one high resolution gallery image and multiple low resolution test images. The test images are captured from surveillance cameras and are collected from multiple sources from the Internet. Since the number of subjects are just seven, in order to mimic a real world scenario, the gallery size is increased to create an \textit{extended} gallery of $1200$ subjects. Images from the Multi-PIE, ND Human Identification Set-B \cite{notredame}, and MEDS\cite{meds} datasets are used for the same. \\ \\ 
\textbf{Protocol:} For all the datasets, a real world matching protocol is followed. For each subject, multiple low resolution images are used as probe images which are matched against the pre-acquired database of high resolution gallery images. Only a single high resolution image per subject is used as gallery. The proposed and comparative algorithms are used to synthesize (or super-resolve) a high resolution image from a given low resolution input. The magnification factor varies from 2 (for probes of $48\times48$) to 12 (for probes of $8\times8$) to match it against the gallery database of size $96\times96$. For all the databases except the SCface, test images are of sizes varying from $8\times8$ to $48\times48$. For the SCface database, predefined protocol is followed and probe resolutions are $24\times24$, $32\times32$, and $48\times48$. Face detection is performed using face co-ordinates (if provided) or using Viola Jones Face Detector \cite{viola} and synthetic downsampling is performed to obtain lower resolutions. 
All the experiments are performed with five times random sub-sampling to ensure consistency. \\
\textbf{Implementation Details:} The SDSR algorithm is trained using the pre-acquired gallery database for each dataset. The regularization constant for sparsity is kept at 0.85. Different dictionaries have different dimensions, based on the input data. For instance, the two-level dictionaries created for SCface dataset contain 100 and 80 atoms in the first and second dictionary respectively. The source code of the algorithm will be made publicly available in order to ensure reproducibility of the proposed approach.

\begin{table*}[]
\centering
\small
\caption{Rank-1 identification accuracies (\%) obtained using Luxand (COTS-II) for cross resolution face recognition. The target resolution is $96\times96$. The algorithms which do not support the required magnification factor are presented as $'-'$. }
\vspace{-10pt}
\label{luxandResults}
\begin{tabular}{|m{1.5em}|m{5em}|m{3.7em}|m{3.5em}|m{3.5em}|m{3.5em}|m{3.5em}|m{3.5em}|m{3.5em}|m{3.5em}|m{4em}|}
\hline
\multirow{2}{*}{} & \textbf{Probe Resolution} & \textbf{Original Image} & \textbf{Bicubic Interp.} & \textbf{Dong \textit{et al.} \cite{fsrcnn}} & \textbf{Kim \textit{et al.} \cite{kwan}} & \textbf{Gu \textit{et al.} \cite{gu_iccv15}} & \textbf{Dong \textit{et al.} \cite{srcnn}} &  \textbf{Peleg \textit{et al.} \cite{peleg}} & \textbf{Yang \textit{et al.} \cite{yang}} & \textbf{Proposed SDSR} \\ 
\hhline {|=|=|=|=|=|=|=|=|=|=|=|}

\multirow{5}{*}{\rotatebox[origin=c]{90}{\parbox[c]{2cm}{\centering \textbf{CMU Multi-PIE}}} }  &
8$\times$8 & 0.0$\pm$0.0 & 0.0$\pm$0.0 & - & - & - & - & - & - & \textbf{82.3$\pm$1.4} \\ \cline{2-11}
& 16$\times$16 & 0.0$\pm$0.0 & 0.0$\pm$0.0 & - & - & - & - & - & - & \textbf{90.5$\pm$1.1} \\ \cline{2-11}
& 24$\times$24 & 0.9$\pm$0.3 & 1.0$\pm$0.3& 2.3$\pm$0.5& 5.9$\pm$0.7& 6.8$\pm$1.4& 6.8$\pm$0.5 & - & - & \textbf{92.1$\pm$1.5} \\ \cline{2-11}
& 32$\times$32 & 11.3$\pm$1.1 & 18.3$\pm$7.1 & 13.5$\pm$0.6 & 28.6$\pm$1.2 & 24.3$\pm$2.3 & 19.4$\pm$1.5 & 17.4$\pm$2.5 & - & \textbf{92.2$\pm$1.6} \\ \cline{2-11}
& 48$\times$48 & 90.2$\pm$0.5 & \textbf{97.9$\pm$0.5} & 96.0$\pm$0.6 & 97.1$\pm$0.7 & 96.6$\pm$0.5 & 96.9$\pm$0.6 & 97.5$\pm$0.5 & 96.2$\pm$0.4 & 91.9$\pm$1.6 \\ 
\hhline {|=|=|=|=|=|=|=|=|=|=|=|}

\multirow{5}{*}{\begin{turn}{90}\textbf{CAS-PEAL}\end{turn}} &
8$\times$8 & 0.0$\pm$0.0 & 0.0$\pm$0.0 & - & - & - & -  & -  & - & \textbf{91.7$\pm$0.9} \\ \cline{2-11}
& 16$\times$16 & 0.0$\pm$0.0 & 0.6$\pm$0.6 & - & - & - & - & - & - & \textbf{93.3$\pm$0.7} \\ \cline{2-11}
& 24$\times$24 & 0.5$\pm$0.4 & 49.3$\pm$1.3 & 2.3$\pm$0.8 & 10.2$\pm$1.1 & 7.3$\pm$1.5 & 5.8$\pm$0.4 & - & - & \textbf{93.7$\pm$1.4} \\ \cline{2-11}
& 32$\times$32 & 11.1$\pm$2.8 & 92.5$\pm$2.0 & 27.9$\pm$1.1 & 34.6$\pm$2.3 & 31.7$\pm$2.3 & 15.2$\pm$3.3 & 28.0$\pm$1.9 & - & \textbf{93.9$\pm$1.3} \\ \cline{2-11}
& 48$\times$48 & 88.1$\pm$0.6 & \textbf{95.4$\pm$1.4} & 85.7$\pm$2.8 & 93.3$\pm$1.4 & 93.3$\pm$1.4 & 91.4$\pm$1.4 & 93.6$\pm$1.9 & 90.8$\pm$1.7 & 93.9$\pm$1.5\\ \cline{2-11}
\hhline {|=|=|=|=|=|=|=|=|=|=|=|}

\multirow{3}{*}{\begin{turn}{90}\textbf{SCface}\end{turn}} &
24$\times$24 & 1.1$\pm$0.2& 1.5$\pm$1.0& 1.2$\pm$0.5 & 0.8$\pm$0.1& 2.2$\pm$0.5 & 1.9$\pm$0.6 & - & - & \textbf{14.7$\pm$2.4} \\ \cline{2-11}
& 32$\times$32 & 2.2$\pm$0.4 & 3.2$\pm$0.4 & 3.5$\pm$0.5 & 2.6$\pm$0.7 & 4.0$\pm$0.7 & 2.6$\pm$1.0 & 2.8$\pm$0.7 & - & \textbf{15.7$\pm$1.3} \\ \cline{2-11}
& 48$\times$48 & 9.7$\pm$1.7& 12.6$\pm$1.7 & 9.6$\pm$1.1 & 11.6$\pm$1.3 & 10.1$\pm$1.8 & 11.7$\pm$2.0 & 11.4$\pm$1.2 & 11.9$\pm$1.0 & \textbf{19.1$\pm$3.4} \\ \hline
\end{tabular}
\vspace{-8pt}
\end{table*}


\section{Results and Analysis}
\label{sec:results}
The proposed algorithm is evaluated with three sets of experiments: (i) face recognition performance with resolution variations, (ii) image quality measure, and (iii) face identification analysis with different dictionary levels. The resolution of the gallery is set to $96\times96$. For the first experiment, the probe resolution varies from $8\times8$ to $48\times48$, while it is fixed to $24\times24$ for the next two experiments.

\subsection{Face Recognition across Resolutions}
For all datasets and resolutions, results are tabulated in Tables \ref{verilookResults} to \ref{realWorld}. The key observations pertaining to these set of experiments are presented below:

\noindent \textbf{$\mathbf{8\times8}$ and $\mathbf{16\times16}$ probe resolutions:} Except bicubic interpolation, none of the existing super resolution or synthesis algorithms used in this comparison support a magnification factor of $12$ (for $8\times8$) or $6$ (for $16\times16$); therefore, the results on these two resolutions are compared with original resolution (when the probe is used as input to COTS as it is, without any resolution enhancement) and bicubic interpolation only. As shown in the third and fourth columns of the two tables, on the CMU Multi-PIE and CAS-PEAL databases, matching with original and bicubic interpolated images results in an accuracy of $\leq 1.1\%$ whereas, the images synthesized using the proposed algorithm provide rank-1 accuracy of $82.6\%$ and $92.8\%$, respectively.

\begin{figure}[!t]
\centering
\includegraphics[width= 3in]{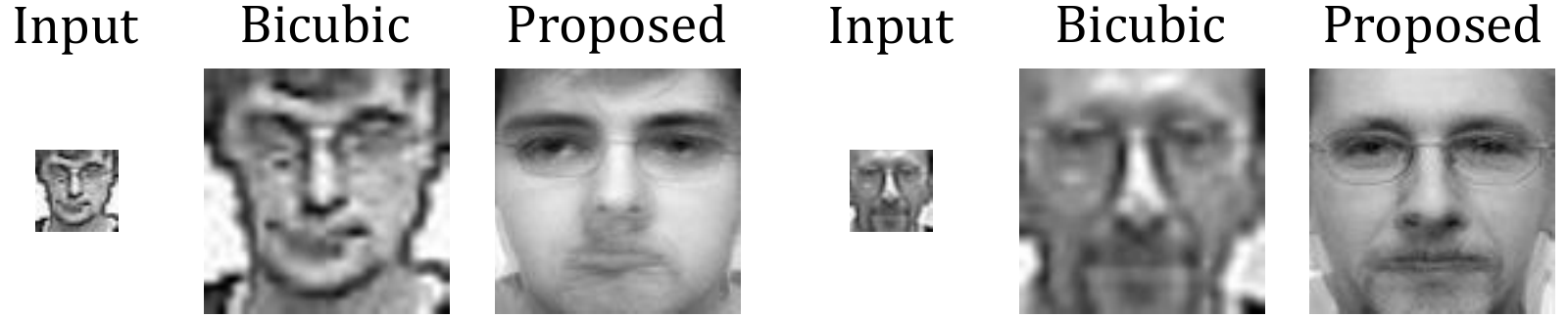}
 \vspace{-5pt}
\caption{Sample images from SCface dataset incorrectly synthesized by the SDSR algorithm for $32\times32$ input. }
 \vspace{-15pt}
\label{bad}
\end{figure}

\noindent \textbf{$\mathbf{24\times24}$ and $\mathbf{32\times32}$ probe resolutions:} As shown in Table \ref{verilookResults}, on CMU Multi-PIE and CAS-PEAL databases with test resolution of $24\times24$ and $32\times32$, the synthesized images obtained using the proposed SDSR algorithm, yield a rank-1 accuracy of $\geq 91.8\%$. Other approaches yield a rank-1 accuracy of less than $20\%$, except bicubic interpolation on  $32\times32$ size which provides rank-1 accuracy of $76.5\%$. As shown in Table \ref{luxandResults}, similar performance trends are observed using COTS-II on the two databases. For SCface, the rank-1 accuracy with SDSR is significantly higher than the existing approaches; however, due to the challenging nature of the database, both commercial matchers provide low rank-1 accuracies. Fig. \ref{bad} presents sample images from the SCface dataset, incorrectly synthesized via the proposed SDSR algorithm. Varying acquisition devices of the training and testing partitions, along with the covariates of pose and illumination creates the problem further challenging.

\noindent \textbf{$\mathbf{48\times48}$ probe resolution:} Using COTS-I, the proposed algorithm achieves improved performance than other techniques, except on the CMU Multi-PIE dataset, where it does not perform as well. On all other databases, the proposed algorithm yields the best results. Upon analyzing both the Tables, it is clear that the proposed algorithm is robust to different recognition systems and performs well without any bias for a specific kind of recognition algorithm.

Another observation is that with COTS-II, images super-resolved using bicubic interpolation yield best results on the first two databases. However, it should be noted that these results are only observed for a magnification factor of $2$ and for images which were synthetically down-sampled. In real world surveillance datasets, such as SCface, the proposed approach performs best with both commercial systems.

\begin{table*}[]
\centering
\small
\caption{Real World Scenarios: Recognition accuracy obtained in top 20\% ranks, against a gallery of 1200 subjects using COTS-I (Verilook) having resolution of 96x96. }
\vspace{-10pt}
\label{realWorld}
\begin{tabular}{|m{4.6em}|m{3.9em}|m{4.8em}|m{4em}|m{4em}|m{4em}|m{4em}|m{4em}|m{4em}|m{4em}|}
\hline
\textbf{Probe Resolution} & \textbf{Original Image} & \textbf{Bicubic Interpolation} & \textbf{Dong \textit{et al.} \cite{fsrcnn}} & \textbf{Kim \textit{et al.} \cite{kwan}} & \textbf{Gu \textit{et al.} \cite{gu_iccv15}} & \textbf{Dong \textit{et al.} \cite{srcnn}} & \textbf{Peleg \textit{et al.} \cite{peleg}} & \textbf{Yang \textit{et al.} \cite{yang}} & {\textbf{Proposed SDSR}} \\ 
\hhline {|=|=|=|=|=|=|=|=|=|=|}

8$\times$8 & 0.0 & 0.0 & - & - & - & - & - & - & \textbf{53.3} \\ 
\hline
 16$\times$16 & 0.0 & 13.3 & - & - & - & - & - & - & \textbf{53.3} \\ 
 \hline
 24$\times$24 & 6.6 & 16.6 & 13.3 & 26.6 & 13.3 & 6.6 & - & - & \textbf{53.3} \\ 
 \hline
 32$\times$32 & 33.3 & 16.6 & 33.3 & 33.3 & 33.3 & 16.6 & 40.0 & - & \textbf{53.3} \\ 
 \hline
 48$\times$48 & 33.3 & 46.6 & 26.6 & 33.3 & 26.6 & 20.0 & 33.3 & 40.0 & \textbf{60.0} \\  
\hline
\end{tabular}
\vspace{-8pt}
\end{table*}

\begin{table*}
\centering
\small
\caption{Average no reference quality measure - BRISQUE \cite{brisque} for probe resolution of $24\times24$ synthesized to $96\times96$, obtained over five folds. A lower value for BRISQUE corresponds to lesser distortions in the image.}
\vspace{-10pt}
\label{tab:quality}
\begin{tabular}{|k{7em}|k{7em}|k{6em}|k{6.1em}|k{6em}|k{6em}|k{7em}|}
\hline
\textbf{Database} & \textbf{Bicubic Interp.} & \textbf{Dong \textit{et al.} \cite{fsrcnn}} & \textbf{Kim \textit{et al.} \cite{kwan}} & \textbf{Gu \textit{et al.} \cite{gu_iccv15}} & \textbf{Dong \textit{et al.} \cite{srcnn}} & {\textbf{Proposed SDSR}} \\ 
\hhline {|=|=|=|=|=|=|=|}
CMU Multi-PIE & {54.8 $\pm$ 0.1} & 28.94 $\pm$ 0.0 & {50.8 $\pm$ 0.1} & {52.8 $\pm$ 0.1} & 48.8 $\pm$ 0.1 &  {\textbf{26.2 $\pm$ 1.3}} \\
\hline 
CAS-PEAL & 60.0 $\pm$ 0.2 & 52.86 $\pm$ 0.0 & 54.3 $\pm$ 0.2 & 56.4 $\pm$ 0.1 & 53.4 $\pm$ 0.1 & \textbf{39.3 $\pm$ 0.3} \\
\hline
SCface & 58.7 $\pm$ 0.1 & 52.86 $\pm$ 0.0 & 53.2 $\pm$ 0.2 & 54.9 $\pm$ 0.1 & 47.2 $\pm$ 0.1 & \textbf{34.2 $\pm$ 0.6} \\
\hline
Real World & {57.5} &28.94 & {54.5} & {54.6} & 49.54 & {\textbf{25.9}} \\ 
\hline
\end{tabular}
\vspace{-10pt}
\end{table*}

\begin{figure}
\centering
\includegraphics[width= 3.2in]{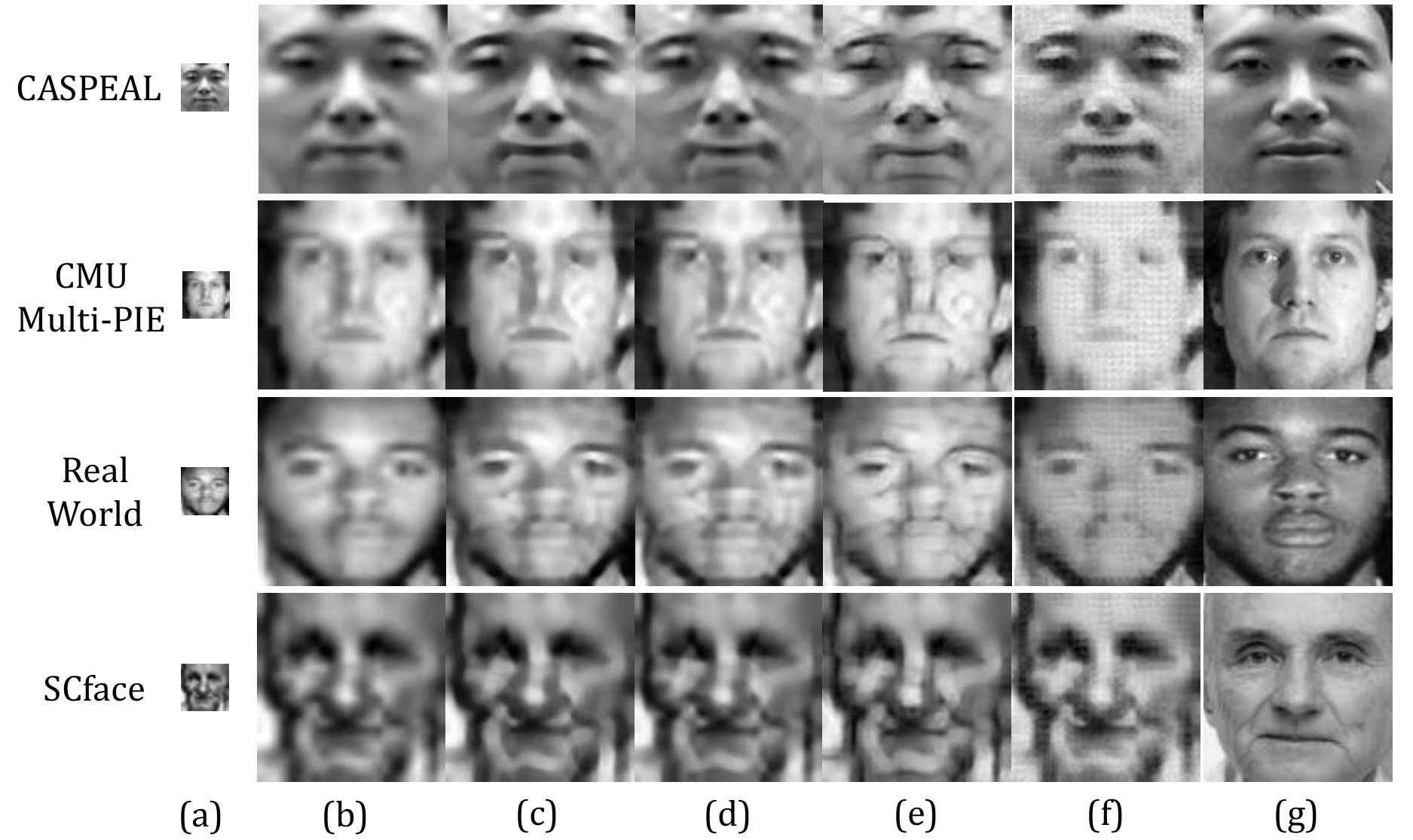}
\vspace{-5pt}
\caption{Probe images of $24\times24$ are super-resolved/synthesized to $96\times96$. (a) corresponds to the original probe, (b)-(f) correspond to different techniques: bicubic interpolation, Kim \textit{et al} \cite{kwan}, Gu \textit{et al.} \cite{gu_iccv15}, Dong \textit{et al.} \cite{srcnn}, Dong \textit{et al.} \cite{fsrcnn}, and the proposed SDSR algorithm.}
\vspace{-15pt}
\label{fig24x24}
\end{figure}

\noindent \textbf{Real World Scenarios Dataset:} Table \ref{realWorld} summarizes the results of COTS-I on Real World Scenarios dataset. Since the gallery contains images from $1200$ subjects, we summarize the results in terms of the identification performance with top 20\% retrieved matches. It is interesting to observe that for all test resolutions, the proposed algorithm significantly outperforms existing approaches. SDSR achieves a 53.3\% identification accuracy on probe resolution of $8\times8$ and an accuracy of 60.0\% for $48\times48$ test resolution.

\noindent \textbf{Cross Dataset Experiments:} The SDSR algorithm was trained on the CMU Multi-PIE dataset, and tested on the SCface dataset for a probe resolution of $24\times24$. 
A rank-1 identification accuracy of 1.62\% (1.92\%) was obtained using COTS-I (COTS-II), whereas a rank-5 identification accuracy of 7.54\% and 9.06\% was obtained respectively. The results showcase that the proposed model is still able to achieve better recognition performance as compared to other techniques. However, the drop in accuracy strengthens our hypothesis that using an identity-aware model for performing synthesis is more beneficial for achieving higher classification performance.  
%
 
\begin{table}
\centering
\small
\caption{Rank-1 accuracies (\%) for varying levels of SDSR algorithm with $24\times24$ probe and $96\times96$ gallery. }
\vspace{-10pt}
\label{proposedLevels}
\begin{tabular}{|m{5em}|m{3.8em}|m{3.8em}|m{3.8em}|m{3.8em}|}
\hline
\multirow{2}{*}{\textbf{Database}} & \multirow{2}{*}{\textbf{COTS}} & \multicolumn{3}{c|}{\textbf{Dictionary Levels}} \\\cline{3-5}
& & \textbf{$k=1$} & \textbf{$k=2$} & \textbf{$k=3$} \\\hline

\textbf{CMU} &Verilook & 91.4 & \textbf{91.8} & \textbf{91.8}\\ \cline {2-5}
\textbf{Multi-PIE} & Luxand & 92.0 & 92.1 & \textbf{92.5}\\ \hline

\multirow{2}{*}{\textbf{CAS-PEAL}} & Verilook & 93.8 & \textbf{95.3} & 93.7 \\\cline {2-5} 
& Luxand & 92.2 & \textbf{93.7}  & 93.6 \\\hline 

\multirow{2}{*}{\textbf{SCface}} & Verilook & 15.0 & 14.7  & \textbf{15.2} \\ \cline {2-5}
& Luxand & \textbf{15.6} & 14.7 & 15.3 \\ \hline
\end{tabular}
\label{tabLevels}
\vspace{-15pt}
\end{table}

\subsection{Quality Analysis}
Fig. \ref{fig24x24} shows examples of synthesized/super-resolved images from multiple databases generated using the proposed and existing algorithms. In this figure, images of $96\times96$ are synthesized from low resolution images of $24\times24$. It can be observed that the output images obtained using existing algorithms (columns (b) - (f)) have artifacts in terms of blockiness and/or blurriness. However, the quality of the images obtained using the proposed algorithm (column (g)) are significantly better than the other algorithms. To compare the visual quality of the outputs, a no reference image quality measure, BRISQUE \cite{brisque} is utilized. Blind/Referenceless Image Spatial QUality Evaluator (BRISQUE) computes the distortion in the image by using the statistics of locally normalized luminance coefficients. It is calculated in the spatial domain and is used to estimate the losses of naturalness in an image. Lower the value, less distorted is an image. From Table \ref{tab:quality}, it can be seen that  images obtained using the proposed SDSR algorithm have a better (lower) BRISQUE score as compared to images generated with existing algorithms; a difference of at least $15$ points is observed in the BRISQUE scores.

\subsection{Effect of Dictionary Levels}
As explained in the algorithm section, synthesis can be performed at different levels of deep dictionary, i.e. with varying values of $k$. This experiment is performed to analyze the effect of different dictionary levels on identification performance. The proposed algorithm is used to synthesize high resolution images ($96\times96$, magnification factor of 4) from input images of size $24\times24$ with varying dictionary levels, i.e. $k=1, 2, 3$.  First level dictionary ($k=1$) is equivalent to shallow dictionary learning, whereas two and three levels correspond to synthesis with deep dictionary learning. Table \ref{tabLevels} reports the rank-1 identification accuracies obtained with the two commercial matchers for four databases. The results show that the proposed approach with $k=2$, generally, yields the best results. In some cases, the proposed approach with $k=3$ yields better results. Generally, abstraction capability of deeper layers and overfitting are two effects in deep learning based approaches. In Table \ref{proposedLevels}, we observe the trade-off between these two. Most of the datasets are moderately sized; therefore, we observe good results in the second layer. In the third layer, \textit{overfitting} offsets the \textit{abstraction}, hence we see none to marginal changes. Further, computational complexity with 3-level deep dictionary features is higher and the improvements in accuracy are not consistent across databases. On the other hand, paired t-test on the results obtained by the shallow dictionary and 2-level deep dictionary demonstrate statistical significance even with a confidence level of 95\% (for Verilook). Specifically, for a single image, synthesis with level-1 dictionary requires 0.42 ms, level-2 requires 0.43 ms, and level-3 requires 0.45 ms.


\section{Conclusion}
The key contribution of this research is a recognition-oriented pre-processing module based on dictionary learning algorithm for synthesizing a high resolution face image from low resolution input. The proposed SDSR algorithm learns the representations of low and high resolution images in a hierarchical manner along with a transformation between the representations of the two. The results are demonstrated on four databases with test image resolutions ranging from $8\times8$ to $48\times48$. Matching these requires generating synthesized high resolution images with a magnification factor of $2$ to $12$. Results computed in terms of both image quality measure and face recognition performance illustrate that the proposed algorithm consistently yields good recognition results. Computationally, the proposed algorithm requires less than 1 millisecond for generating a synthesized high resolution image which further showcases the efficacy and usability of the algorithm for low resolution face recognition applications. 

{\small
\bibliographystyle{ieee}
\bibliography{bibFile}

\begin{thebibliography}{10}\itemsep=-1pt

\bibitem{luxand}
Luxand.
\newblock https://www.luxand.com.

\bibitem{verilook}
Verilook.
\newblock http://www.neurotechnology.com/verilook.html.

\bibitem{baker2000}
S.~Baker and T.~Kanade.
\newblock Hallucinating faces.
\newblock In {\em IEEE International Conference on Automatic Face and Gesture
  Recognition}, FG '00, pages 83--, 2000.

\bibitem{himanshu}
H.~S. Bhatt, R.~Singh, M.~Vatsa, and N.~K. Ratha.
\newblock Improving cross-resolution face matching using ensemble-based
  co-transfer learning.
\newblock {\em IEEE Transactions on Image Processing}, 23(12):5654--5669,
  December 2014.

\bibitem{pixel17Iccv}
R.~Dahl, M.~Norouzi, and J.~Shlens.
\newblock Pixel recursive super resolution.
\newblock In {\em IEEE International Conference on Computer Vision}, 2017.

\bibitem{srcnn}
C.~Dong, C.~C. Loy, K.~He, and X.~Tang.
\newblock Image super-resolution using deep convolutional networks.
\newblock {\em IEEE Transactions on Pattern Analysis and Machine Intelligence},
  38(2):295--307, 2016.

\bibitem{fsrcnn}
C.~Dong, C.~C. Loy, and X.~Tang.
\newblock Accelerating the super-resolution convolutional neural network.
\newblock In {\em European Conference on Computer Vision}, pages 391--407.
  Springer, 2016.

\bibitem{notredame}
P.~J. Flynn, K.~W. Bowyer, and P.~J. Phillips.
\newblock Assessment of time dependency in face recognition: An initial study.
\newblock In {\em International Conference on Audio-and Video-Based Biometric
  Person Authentication}, pages 44--51, 2003.

\bibitem{meds}
A.~Founds, N.~Orlans, G.~Whiddon, and C.~Watson.
\newblock {NIST} special database 32-multiple encounter dataset {II}
  ({MEDSII}).
\newblock {\em National Institute of Standards and Technology, Tech. Rep},
  2011.

\bibitem{caspeal}
W.~Gao, B.~Cao, S.~Shan, X.~Chen, D.~Zhou, X.~Zhang, and D.~Zhao.
\newblock {The} {CAS-PEAL} large-scale chinese face database and baseline
  evaluations.
\newblock {\em IEEE Transactions on Systems, Man, and Cybernetics - Part A:
  Systems and Humans}, 38(1):149--161, January 2008.

\bibitem{scface}
M.~Grgic, K.~Delac, and S.~Grgic.
\newblock Scface - surveillance cameras face database.
\newblock {\em Multimedia Tools Application}, 51(3):863--879, February 2011.

\bibitem{mpie}
R.~Gross, I.~Matthews, J.~Cohn, T.~Kanade, and S.~Baker.
\newblock Multi-{PIE}.
\newblock {\em Image Vision Computing}, 28(5):807--813, May 2010.

\bibitem{gu_iccv15}
S.~Gu, W.~Zuo, Q.~Xie, D.~Meng, X.~Feng, and L.~Zhang.
\newblock Convolutional sparse coding for image super-resolution.
\newblock In {\em IEEE International Conference on Computer Vision}, December
  2015.

\bibitem{jian2015}
M.~Jian and K.~M. Lam.
\newblock Simultaneous hallucination and recognition of low-resolution faces
  based on singular value decomposition.
\newblock {\em IEEE Transactions on Circuits and Systems for Video Technology},
  25(11):1761--1772, November 2015.

\bibitem{kwan}
K.~I. Kim and Y.~Kwon.
\newblock Single-image super-resolution using sparse regression and natural
  image prior.
\newblock {\em IEEE Transactions on Pattern Analysis and Machine Intelligence},
  32(6):1127--1133, June 2010.

\bibitem{ledig17Cvpr}
C.~Ledig, L.~Theis, F.~Huszar, J.~Caballero, A.~Cunningham, A.~Acosta,
  A.~Aitken, A.~Tejani, J.~Totz, Z.~Wang, and W.~Shi.
\newblock Photo-realistic single image super-resolution using a generative
  adversarial network.
\newblock In {\em IEEE Conference on Computer Vision and Pattern Recognition},
  2017.

\bibitem{dictLearning}
H.~Lee, A.~Battle, R.~Raina, and A.~Y. Ng.
\newblock Efficient sparse coding algorithms.
\newblock In {\em Advances in Neural Information Processing Systems}, pages
  801--808. 2007.

\bibitem{brisque}
A.~Mittal, A.~K. Moorthy, and A.~C. Bovik.
\newblock No-reference image quality assessment in the spatial domain.
\newblock {\em IEEE Transactions on Image Processing}, 21(12):4695--4708, 2012.

\bibitem{mundunuri16}
S.~P. Mudunuri and S.~Biswas.
\newblock Low resolution face recognition across variations in pose and
  illumination.
\newblock {\em IEEE Transactions on Pattern Analysis and Machine Intelligence},
  38(5):1034--1040, 2016.

\bibitem{sparseFilt}
J.~Ngiam, Z.~Chen, S.~A. Bhaskar, P.~W. Koh, and A.~Y. Ng.
\newblock Sparse filtering.
\newblock In {\em Advances in Neural Information Processing Systems}, pages
  1125--1133. 2011.

\bibitem{peleg}
T.~Peleg and M.~Elad.
\newblock A statistical prediction model based on sparse representations for
  single image super-resolution.
\newblock {\em IEEE Transactions on Image Processing}, 23(6):2569--2582, June
  2014.

\bibitem{bayesianSR}
G.~Polatkan, M.~Zhou, L.~Carin, D.~Blei, and I.~Daubechies.
\newblock A bayesian nonparametric approach to image super-resolution.
\newblock {\em IEEE Transactions on Pattern Analysis and Machine Intelligence},
  37(2):346--358, 2015.

\bibitem{doubleSparsity}
R.~Rubinstein, M.~Zibulevsky, and M.~Elad.
\newblock Double sparsity: Learning sparse dictionaries for sparse signal
  approximation.
\newblock {\em IEEE Transactions on Signal Processing}, 58(3):1553--1564, March
  2010.

\bibitem{deepDict}
S.~Tariyal, A.~Majumdar, R.~Singh, and M.~Vatsa.
\newblock Greedy deep dictionary learning.
\newblock {\em CoRR}, abs/1602.00203, 2016.

\bibitem{multiLevel}
J.~J. Thiagarajan, K.~N. Ramamurthy, and A.~Spanias.
\newblock Multilevel dictionary learning for sparse representation of images.
\newblock In {\em 2011 Digital Signal Processing and Signal Processing
  Education Meeting (DSP/SPE)}, pages 271--276, 2011.

\bibitem{tong17Iccv}
T.~Tong, G.~Li, X.~Liu, and Q.~Gao.
\newblock Image super-resolution using dense skip connections.
\newblock In {\em IEEE International Conference on Computer Vision}, 2017.

\bibitem{viola}
P.~Viola and M.~J. Jones.
\newblock Robust real-time face detection.
\newblock {\em International Journal Computer Vision}, 57(2):137--154, 2004.

\bibitem{wang14Rev}
N.~Wang, D.~Tao, X.~Gao, X.~Li, and J.~Li.
\newblock A comprehensive survey to face hallucination.
\newblock {\em International journal of computer vision}, 106(1):9--30, 2014.

\bibitem{scdl}
S.~Wang, L.~Zhang, Y.~Liang, and Q.~Pan.
\newblock Semi-coupled dictionary learning with applications to image
  super-resolution and photo-sketch synthesis.
\newblock In {\em IEEE Conference on Computer Vision and Pattern Recognition},
  pages 2216--2223, 2012.

\bibitem{wang16CVPR}
Z.~Wang, S.~Chang, Y.~Yang, D.~Liu, and T.~S. Huang.
\newblock Studying very low resolution recognition using deep networks.
\newblock In {\em The IEEE Conference on Computer Vision and Pattern
  Recognition}, June 2016.

\bibitem{Wang2013}
Z.~Wang, Z.~Miao, Q.~M. Jonathan~Wu, Y.~Wan, and Z.~Tang.
\newblock Low-resolution face recognition: a review.
\newblock {\em The Visual Computer}, 30(4):359--386, 2013.

\bibitem{yang}
J.~Yang, J.~Wright, T.~S. Huang, and Y.~Ma.
\newblock Image super-resolution via sparse representation.
\newblock {\em IEEE Transactions on Image Processing}, 19(11):2861--2873,
  November 2010.

\bibitem{icb}
M.~C. Yang, C.~P. Wei, Y.~R. Yeh, and Y.~C.~F. Wang.
\newblock Recognition at a long distance: Very low resolution face recognition
  and hallucination.
\newblock In {\em International Conference on Biometrics}, pages 237--242, May
  2015.

\end{thebibliography}
}

\end{document}